\documentclass[APA,STIX2COL]{WileyNJD-v2}

\usepackage[natbibapa]{apacite}%
\usepackage{balance}
\usepackage{subcaption}
\usepackage{xcolor}
\definecolor{darkred}{RGB}{139,0,0}

\usepackage[linesnumbered,ruled,algo2e]{algorithm2e}
\usepackage{algpseudocode}
\usepackage{amsmath}
\usepackage{multirow}
\usepackage{makecell}
\usepackage{float}
\usepackage{tabularray}
\usepackage{array,ragged2e}
\usepackage{xcolor}

\SetCommentSty{mycommfont}
\newlength\algowd

\usepackage{textcomp}
\usepackage{listings}
\usepackage{xcolor}
\usepackage{float}
\definecolor{codebg}{gray}{0.95}
\makeatletter
\def\NAT@aysep{,}
\makeatother

\articletype{ORIGINAL ARTICLE}%

\received{NA}
\revised{NA}
\accepted{NA}

\begin{document}

\title{Training-free retrieval-augmented generation with reinforced reasoning for flood damage nowcasting}

\author[1]{Lipai Huang}
\author[2]{Kai Yin*}
\author[3]{Chia-Fu Liu}
\author[1,4]{Ali Mostafavi}

\address[1]{\orgdiv{Urban Resilience.AI Lab, Zachry Department of Civil and Environmental Engineering}, \orgname{Texas A\&M University}, \orgaddress{\state{Texas}, \country{USA}}}
\address[2]{\orgdiv{Department of Computer Science and Engineering}, \orgname{Texas A\&M University}, \orgaddress{\state{Texas}, \country{USA}}}
\address[3]{\orgdiv{Department of Civil, Environmental and Architectural Engineering}, \orgname{University of Kansas}, \orgaddress{\state{Kansas}, \country{USA}}}
\address[4]{\orgdiv{Institute for a Disaster Resilient Texas}, \orgname{Texas A\&M University}, \orgaddress{\state{Texas}, \country{USA}}}

\corres{Kai Yin, Department of Computer Science and Engineering, Texas A\&M University, College Station, TX, USA\\
\email{\href{kai_yin@tamu.edu}{kai\_yin@tamu.edu}}\\}

\abstract[Abstract]{
We propose \textit{R2RAG-Flood}, a training-free retrieval-augmented generation framework for flood damage nowcasting with reinforced reasoning. The framework builds a reasoning-centric knowledge base from labeled tabular records, where each sample includes structured predictors, a compact text-mode summary, and a model-generated reasoning trajectory. During inference, the target prompt is augmented with geographically local neighbors and selected free-shots to support case-based reasoning without task-specific fine-tuning. A two-stage procedure first determines damage occurrence and then refines severity within a three-level Property Damage Extent (PDE) classification, followed by a conservative downgrade check for weakly supported over-severe outputs. In a Hurricane Harvey case study in Harris County, Texas, the supervised tabular baseline achieves 0.714 overall accuracy and 0.859 accuracy on the damaged classes (medium and high PDE). Across seven LLM backbones, \textit{R2RAG-Flood} achieves 0.613--0.668 overall accuracy and 0.757--0.896 accuracy on the damaged classes while providing a structured rationale for each prediction. Under the severity-per-cost metric used in this study, lighter \textit{R2RAG-Flood} variants are more cost-efficient than the supervised baseline and larger LLM backbones. These results demonstrate the feasibility of a reasoning-centric, training-free pipeline for flood damage nowcasting in a realistic case-study setting.
}


\maketitle

\section{INTRODUCTION}\label{sec:1}

\begin{figure*}[htbp]
    \centering
    \includegraphics[width=0.9\textwidth]{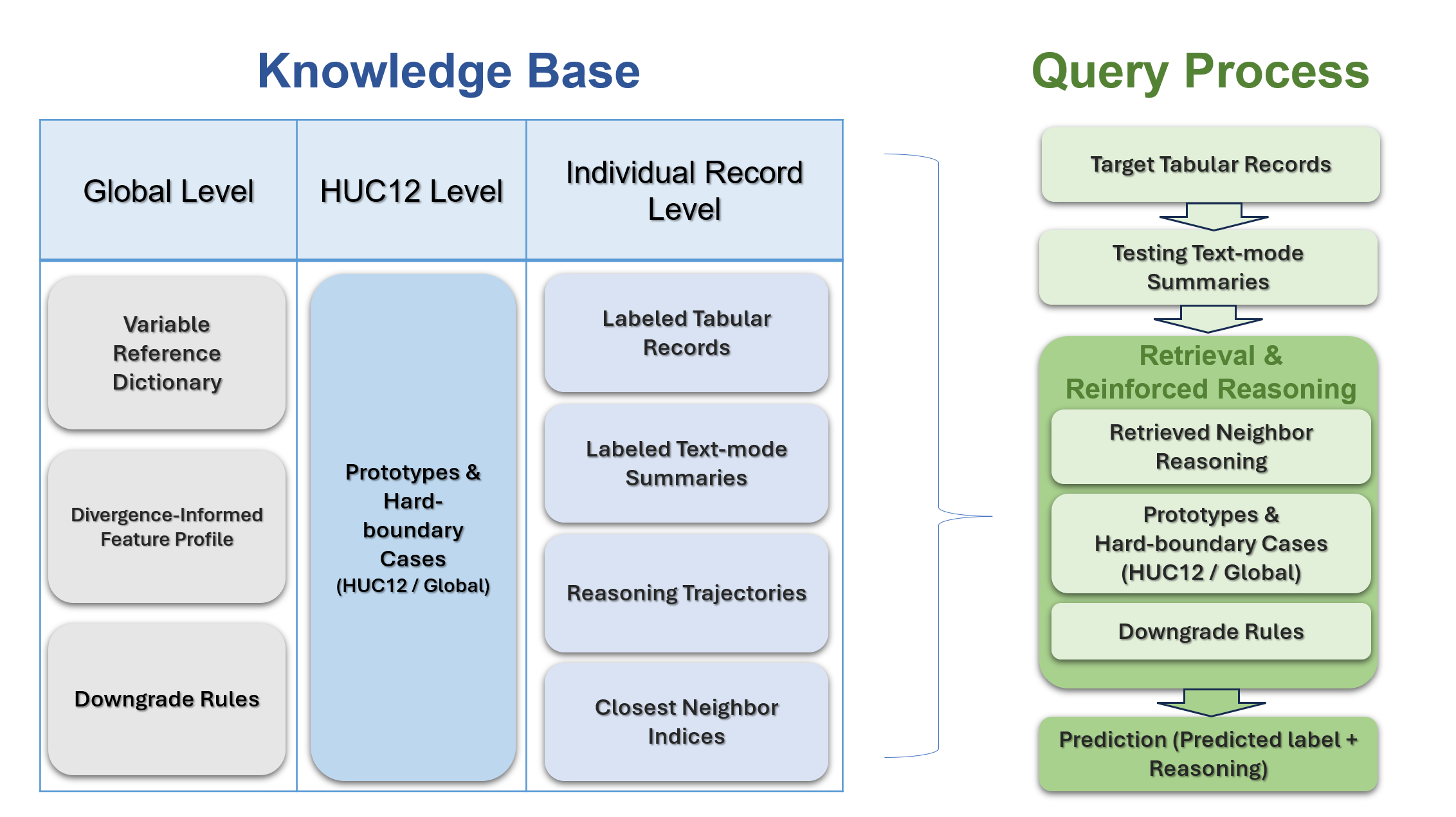}
    \caption{Schematic of \textit{R2RAG-Flood}. A three-level reasoning-centric knowledge base is constructed from global references, HUC12-level prototypes and hard-boundary cases, and individual labeled records. During inference, each target tabular record is converted into a text-mode summary and evaluated through retrieval and reinforced reasoning, producing both a PDE classification and a structured rationale.}
    \label{fig:main}
\end{figure*}

Flood hazards impose major physical and economic burdens on communities worldwide, with U.S. flood-related losses averaging about 40~billion dollars annually over the past four decades and likely to increase as events become more frequent and intense \citep{yildirim2022flood,slater2016recent,musselman2018projected}. Recent studies have advanced computational flood assessment from multiple directions, including predictive flood warning from sensor networks \citep{dong2021hybrid}, post-disaster building and infrastructure assessment from image-based intelligence \citep{braik2024automated,ho2025integrated}, community-level flood functionality and risk analysis \citep{nofal2024community,ewis2025leveraging}, physics-informed flood forecasting \citep{taghizadeh2025interpretable}, and supervised tabular nowcasting from claim and geospatial data \citep{liu2024flooddamagecast}. More broadly, supervised classification research has continued to develop efficient learning schemes, including neural dynamic classification and fast-learning models such as FEMa \citep{rafiei2017new,pereira2020fema}. These studies demonstrate the value of data-driven flood intelligence, but practical deployment often still depends on substantial task-specific preparation, including dataset assembly, feature engineering, model training, and imbalance handling. Adapting such workflows to a new setting can therefore remain costly and slow.

This study examines whether retrieval and prompting can provide a practical alternative to repeated retraining for flood damage nowcasting. We propose \textit{R2RAG-Flood}, a retrieval-augmented generation framework with reinforced reasoning for post-storm Property Damage Extent (PDE) prediction. In this paper, ``training-free'' means that the LLM backbone is used without task-specific gradient-based training or fine-tuning at deployment time, although the pipeline still requires offline construction of a knowledge base from labeled records. Unlike document-oriented RAG tasks, flood damage nowcasting is driven by structured tabular predictors, local hydrologic comparability, and boundary-sensitive ordinal decisions. These characteristics motivate a flood-specific retrieval and reasoning pipeline built around labeled flood cases rather than a generic document RAG design.

Claim-based flood damage data are also highly imbalanced. Reporting is incomplete for many affected locations, and observed claims often correspond to substantial losses, which makes direct regression on claim amounts sensitive to noise and extremes \citep{mobley2021quantification}. We therefore formulate PDE nowcasting as an ordinal classification problem. PDE labels are defined using domain-informed, distribution-based thresholds that provide stable severity levels for retrieval and reasoning. In supervised learning, imbalance is often addressed through class reweighting and synthetic augmentation, including conditional tabular synthesis methods such as CTGAN \citep{xu2019modeling,SDV}. By contrast, \textit{R2RAG-Flood} relies on retrieval and reasoning over labeled cases, so synthetic augmentation is not a central component of the framework.

Figure~\ref{fig:main} summarizes the proposed workflow. \textit{R2RAG-Flood} organizes prior labeled information into a three-level knowledge base comprising global references, HUC12-level example libraries, and record-level entries. The components of the knowledge base are derived from tabular records and domain-specific settings. An offline divergence analysis based on Jensen--Shannon divergence and the Kolmogorov--Smirnov statistic \citep{menendez1997jensen,massey1951kolmogorov} is used to prioritize features, support free-shot construction, and define conservative post-check rules. At inference time, the target record is converted into a text-mode summary and evaluated using retrieved neighbors, selected boundary examples, and downgrade checks drawn from the knowledge base. The model outputs both a PDE label and the corresponding rationale. To preserve hydrologic comparability, retrieval is performed within 12-digit Hydrologic Unit Code (HUC12) watershed groups when sufficient local examples are available, with a global fallback in sparse areas \citep{luukkonen2023public}.

Our goal is to assess the practical viability of this pipeline in a realistic case-study setting rather than to claim broad geographic generalization from a single event. Using Hurricane Harvey in Harris County, Texas, we examine whether a no-fine-tuning retrieval pipeline can produce useful PDE predictions, generate structured reasoning, and achieve competitive severity-per-cost efficiency relative to a supervised tabular baseline across multiple LLM backbones.

The main contributions of this work are as follows:
\begin{itemize}
    \item[1] We introduce \textit{R2RAG-Flood}, a training-free retrieval-augmented generation framework for flood damage nowcasting that uses a three-level reasoning-centric knowledge base and avoids task-specific fine-tuning of the LLM backbone at deployment time.
    \item[2] We develop a flood-specific inference pipeline that combines geographically local neighbor evidence, HUC12/global prototypes and hard-boundary cases, and a conservative downgrade check to support more reliable ordinal PDE decisions.
    \item[3] We provide a Hurricane Harvey case-study evaluation in Harris County showing that the proposed pipeline can generate useful PDE predictions and structured rationales without task-specific fine-tuning, while offering competitive severity-per-cost efficiency across multiple LLM backbones.
\end{itemize}
\section{REASONING-CENTRIC KNOWLEDGE BASE AND RETRIEVAL FRAMEWORK}\label{sec:2}

\textit{R2RAG-Flood} organizes labeled flood records into a reasoning-centric knowledge base and uses retrieval-augmented prompting for PDE prediction. Here, ``training-free'' means that the LLM backbone is used without task-specific gradient-based training or fine-tuning; it does not mean that the framework has no offline preprocessing. The offline components are used to standardize feature semantics, convert tabular records into compact text-mode summaries, generate reasoning trajectories for labeled examples, and construct reusable free-shot libraries. At prediction time, the target record is converted into the same text-mode format and prompted together with retrieved local neighbors and, when needed, selected free-shots from the prebuilt knowledge base. Throughout this section, all indexed records, neighbor pools, and free-shot libraries are constructed from the training split only; no test labels are indexed or retrieved during inference. The following subsections describe the shared feature dictionary, divergence-based feature profiling, prompt-based text-mode and reasoning generation, free-shot construction, retrieval-conditioned prediction, and the conditional downgrade mechanism.

\subsection{Variable Reference Dictionary and Feature Families}

All predictors used in this study are compiled into a unified Variable Reference Dictionary so that the same feature semantics are used in knowledge-base construction and prediction prompts. The dictionary covers built-environment, topographic, hydrologic, and event-specific predictors together with the PDE label. Each entry stores the feature abbreviation, full name, physical unit, and a brief description. This dictionary is used as metadata when tabular inputs are converted into text-mode summaries, helping the LLM interpret numerical values more consistently.

\begin{table*}[h]
\centering
\footnotesize
\setlength{\tabcolsep}{3pt}
\caption{Prompt template for text-mode generation, applied to both labeled and target records. \texttt{ORDERED\_FEATURES} is the feature list ordered by divergence analysis in Section~\ref{sec:distri}. \texttt{LEGEND} is a compact string of variable meanings and units. \texttt{INPUT\_JSONL} is the JSON Lines input with one sample per line.}
\label{tab:prompt_textmode}
\begin{tabularx}{\linewidth}{@{}l X@{}}
\toprule
\textbf{Role} & \textbf{Template Prompt} \\
\midrule
System &
Convert tabular rows into a concise human-readable
\textcolor{darkgray}{\texttt{text\_mode}}.
Output STRICT JSONL, one line per row:
\textcolor{darkgray}{\texttt{\{"row\_id":<int>, "text\_mode":"<string>"\}}}.
No extra text. \\[2pt]
User &
Write a $\le$120-word paragraph that only paraphrases feature values.
Follow order: [\textcolor{darkgray}{\texttt{ORDERED\_FEATURES}}].
Do not infer risk or predict. Skip NULL or trivial zeros. Use units from the legend when useful.
Keep numbers compact. Input legend: [\textcolor{darkgray}{\texttt{LEGEND}}].
Input JSONL: [\textcolor{darkgray}{\texttt{INPUT\_JSONL}}]. \\
\bottomrule
\end{tabularx}
\end{table*}

\subsection{Feature Distribution Divergence Analysis}\label{sec:distri}

To support both damage occurrence and damage severity decisions, we quantify how strongly each predictor separates PDE levels through a feature-distribution divergence analysis. The resulting scores serve as a design signal in \textit{R2RAG-Flood}: they determine the feature order used in text-mode generation and define the task-specific feature profile later used for free-shot construction and downgrade support.

For each feature $f_i$, we evaluate distributional separation under two comparisons. The first compares the low PDE class against the union of the damage classes (medium and high), targeting the occurrence boundary. The second compares the medium and high PDE classes, targeting the severity boundary. For each comparison, we compute the Kolmogorov--Smirnov (KS) statistic \citep{massey1951kolmogorov} and the Jensen--Shannon (JS) divergence \citep{menendez1997jensen}:
\begin{equation}\label{eq:KS}
D^{(i)}_{\mathrm{KS}} = \sup_{x}\big|F^{(i)}_{a}(x) - F^{(i)}_{b}(x)\big|
\end{equation}
where $F^{(i)}_{a}$ and $F^{(i)}_{b}$ are the cumulative distributions of feature $f_i$ for groups $a$ and $b$. Let $P^{(i)}_{a}$ and $P^{(i)}_{b}$ be the corresponding probability densities. The JS divergence is
\begin{subequations}
\begin{equation}\label{eq:JS}
D^{(i)}_{\mathrm{JS}} = \tfrac{1}{2} D_{\mathrm{KL}}\!\left(P^{(i)}_{a}\|M^{(i)}\right)
+ \tfrac{1}{2} D_{\mathrm{KL}}\!\left(P^{(i)}_{b}\|M^{(i)}\right)
\end{equation}
\begin{equation}
M^{(i)} = \tfrac{1}{2}\left(P^{(i)}_{a} + P^{(i)}_{b}\right)
\end{equation}
\end{subequations}
where $D_{\mathrm{KL}}$ is the Kullback--Leibler divergence. We then define a composite divergence score as
\begin{equation}\label{eq:score}
\mathrm{Score}^{(i)} = W_{\mathrm{JS}}\,D^{(i)}_{\mathrm{JS}} + W_{\mathrm{KS}}\,D^{(i)}_{\mathrm{KS}}.
\end{equation}
In this study, we use $W_{\mathrm{JS}}=0.7$ and $W_{\mathrm{KS}}=0.3$, placing greater emphasis on overall distributional-shape difference while retaining KS as a complementary separation signal. This weighting is a practical heuristic for the current prediction setting.

We compute Eq.~\ref{eq:score} for both comparisons and obtain ordered feature lists for occurrence and severity. Because text-mode generation is length-constrained, predictors with higher divergence scores are placed earlier so that the most boundary-relevant evidence appears first in each summary. The resulting rankings are reported in Table~\ref{tab:divergence_scores}. For the current setting, the retained task-specific feature subset is defined as the top five features in the corresponding ranking for both occurrence-related and severity-related use. These retained subsets are used later in text-mode generation, free-shot selection, and rationale evaluation. The same divergence profile further provides supporting signals for the downgrade check in Section~\ref{sec:downgrade}.

\subsection{Text-Mode Generation from Tabular Records}\label{sec:prompt_text_mode}

Inspired by table-to-text work such as ToTTo \citep{parikh2020totto}, we convert JSONized tabular predictors into compact text-mode summaries. The prompt template in Table~\ref{tab:prompt_textmode} is applied uniformly to both labeled examples and prediction targets. Each text mode is limited to at most 120 words and only paraphrases feature values; it does not infer risk or predict damage. The 120-word cap is used to keep the summary concise while still covering the main high-ranking predictors.

To enforce a consistent structure, the prompt provides an ordered feature list from Section~\ref{sec:distri} and a legend of variable meanings and units from the Variable Reference Dictionary. The feature order follows the divergence rankings in Table~\ref{tab:divergence_scores}. Because each text mode is capped at 120 words, the retained task-specific high-ranking features are placed first, and lower-ranked predictors are omitted once the length limit is reached. This keeps the summaries compact while preserving the most class-separating evidence.

\subsection{LLM Reasoning Trajectories for Labeled Examples}\label{sec:reasoning}

For labeled tabular records, we prompt the LLM to generate reasoning trajectories that explain the observed PDE label given the text-mode summary. We follow the DeepSeek-R1 style \citep{guo2025deepseek} and wrap each trajectory in a \textcolor{darkgray}{\texttt{<think>...</think>}} block followed by a \textcolor{darkgray}{\texttt{<answer>...</answer>}} block. The \textcolor{darkgray}{\texttt{<answer>}} content is constrained to exactly match the known PDE label. These trajectories are stored as auxiliary retrieved context in the knowledge base; they are not treated as ground-truth expert explanations.

The reasoning prompt in Table~\ref{tab:prompt_reasoning} follows a two-stage structure. It first resolves occurrence (low PDE versus damage), then resolves severity (medium versus high) when the outcome is nonzero. This structure is used to align the prompt with the ordinal PDE task; it is a task design choice rather than a separately validated superiority claim over all alternative prompt formats.

The prompt also requires exactly one sentence beginning with ``Despite.'' This is a formatting device used to encourage the model to state an explicit conflict between supporting and opposing evidence. Its purpose is to make retrieved rationales more structured and easier to compare across labeled cases.

\begin{table*}[ht]
\centering
\footnotesize
\setlength{\tabcolsep}{3pt}
\caption{Prompt template for knowledge-base reasoning. \texttt{INPUT\_JSONL} is the JSON Lines input with one sample per line (row ID, text mode, ground-truth target, and HUC12 ID).}
\label{tab:prompt_reasoning}
\begin{tabularx}{\linewidth}{@{}l X@{}}
\toprule
\textbf{Role} & \textbf{Template Prompt} \\
\midrule
System &
You are an expert flood risk analyst. \\
& \textbf{Task}: For each input item, output STRICT JSONL with exactly one line:
\textcolor{darkgray}{\texttt{\{"row\_id":<int>, "r1":"<think>...</think><answer>...</answer>"\}}}. \\
& \textbf{Format rules}: \textcolor{darkgray}{\texttt{r1}} contains exactly one \textcolor{darkgray}{\texttt{<think>}} block immediately followed by one \textcolor{darkgray}{\texttt{<answer>}} block. Tag names are lowercase. Nothing before \textcolor{darkgray}{\texttt{<think>}} or after \textcolor{darkgray}{\texttt{</answer>}}. The content of \textcolor{darkgray}{\texttt{<answer>}} is exactly one of \textcolor{darkgray}{\texttt{"0"}}, \textcolor{darkgray}{\texttt{"1"}}, \textcolor{darkgray}{\texttt{"2"}}. Output is valid single-line JSON with no trailing commas. \\
& \textbf{Reasoning rules}: use only the provided \textcolor{darkgray}{\texttt{text\_mode}} and the given \textcolor{darkgray}{\texttt{ground\_truth}}. Write about 200 words. Apply a two-stage structure: first occurrence (class \textcolor{darkgray}{\texttt{0}} versus classes \textcolor{darkgray}{\texttt{1}} or \textcolor{darkgray}{\texttt{2}}), then severity (class \textcolor{darkgray}{\texttt{1}} versus class \textcolor{darkgray}{\texttt{2}} when the outcome is nonzero). For occurrence, focus on structural, exposure, and history cues such as stream proximity, foundation height, building density or imperviousness, prior claims, and elevation. Rainfall intensity alone must not force a nonzero occurrence. Imperviousness or prior claims can support nonzero, but they do not overrule strong protection cues such as higher elevation or a raised foundation if \textcolor{darkgray}{\texttt{text\_mode}} frames them as protective. For severity, discuss rainfall intensity, duration and accumulation, local drainage such as HAND or elevation, and vulnerability such as FAR or low clearance. Use the exact phrases \textcolor{darkgray}{\texttt{"occurrence resolves to ..."}} and \textcolor{darkgray}{\texttt{"severity resolves to ..."}} when applicable. Include exactly one sentence that begins with \textcolor{darkgray}{\texttt{"Despite"}} to resolve conflicts. End \textcolor{darkgray}{\texttt{<think>}} with \textcolor{darkgray}{\texttt{"Based on these factors, it is reasonable to claim PDE\_category is X."}}. The content of \textcolor{darkgray}{\texttt{<answer>}} must equal the provided \textcolor{darkgray}{\texttt{ground\_truth}}. If \textcolor{darkgray}{\texttt{<answer>}} is \textcolor{darkgray}{\texttt{"0"}}, do not claim \textcolor{darkgray}{\texttt{"severity resolves to ..."}}; you may state a conditional sentence about severity. \\[2pt]
User &
For each item you will receive \textcolor{darkgray}{\texttt{text\_mode}}, \textcolor{darkgray}{\texttt{ground\_truth}} where \textcolor{darkgray}{\texttt{0=L}}, \textcolor{darkgray}{\texttt{1=M}}, \textcolor{darkgray}{\texttt{2=H}}, and \textcolor{darkgray}{\texttt{huc12}}. \\
& Write a single \textcolor{darkgray}{\texttt{<think>...</think>}} block followed immediately by a single \textcolor{darkgray}{\texttt{<answer>...</answer>}} block, following the rules above. Include one conflict sentence that starts with \textcolor{darkgray}{\texttt{"Despite"}}. Conclude the \textcolor{darkgray}{\texttt{<think>}} block with \textcolor{darkgray}{\texttt{"Based on these factors, it is reasonable to claim PDE\_category is X."}}. Ensure that \textcolor{darkgray}{\texttt{<answer>}} exactly matches \textcolor{darkgray}{\texttt{ground\_truth}} (\textcolor{darkgray}{\texttt{"0"}}, \textcolor{darkgray}{\texttt{"1"}}, or \textcolor{darkgray}{\texttt{"2"}}). \\
& Finally, output STRICT JSONL with no extra lines or commentary:
\textcolor{darkgray}{\texttt{\{"row\_id":<int>, "r1":"<think>...</think><answer>...</answer>"\}}}. \\
& Inputs: [\textcolor{darkgray}{\texttt{INPUT\_JSONL}}] \\
\bottomrule
\end{tabularx}
\end{table*}

\subsection{Free-Shot Construction}\label{sec:freeshot}

In \textit{R2RAG-Flood}, free-shots are not learned through parameter updates. They are selected offline from labeled examples and reused as prompt exemplars during inference. For each labeled record, the knowledge base stores the tuple \{\textcolor{darkgray}{\texttt{record}}, \textcolor{darkgray}{\texttt{text\_mode}}, \textcolor{darkgray}{\texttt{reasoning}}\}. Based on these stored entries, we construct a free-shot library for each HUC12 group. When a HUC12 contains fewer than 100 labeled training records, the local sample is treated as too sparse for stable free-shot construction in this case study, and we fall back to a global library built from all training records.

Selection is guided by the divergence-informed feature profile from Section~\ref{sec:distri}. Let $\mathcal{F}$ denote the retained task-specific feature subset defined as the top five features in the corresponding ranking in Table~\ref{tab:divergence_scores} for both occurrence-related and severity-related selection. Thus, $\mathcal{F}$ is task-specific rather than all 14 predictors used uniformly. Let $w_j \in (0,1)$ be the normalized cue weight for feature $j \in \mathcal{F}$. For a record $r$ and an ordinal PDE level $\ell\in\{\mathrm{Low},\mathrm{Medium},\mathrm{High}\}$, we compute per-feature standardized distances
\begin{equation}
z_{r,j}^{(\ell)} = \frac{\lvert x_{r,j} - \mu_{\ell,j} \rvert}{\sigma_{\ell,j} + \epsilon}
\end{equation}
where $x_{r,j}$ is the value of feature $j$ for record $r$, and $\mu_{\ell,j}$ and $\sigma_{\ell,j}$ are the mean and standard deviation of feature $j$ among records at level $\ell$ within the same HUC12; under global fallback, these statistics are computed from the global training pool. The term $\epsilon=10^{-6}$ is included only for numerical stability when the within-level standard deviation is extremely small. We then define a divergence-weighted z-distance
\begin{equation}
d_{r}^{(\ell)} = \sum_{j \in \mathcal{F}} w_j \, z_{r,j}^{(\ell)}. \label{eq:w-z}
\end{equation}

Within each HUC12, for each PDE level $\ell$, we select two prototypes as the labeled records with the smallest $d_{r}^{(\ell)}$. These cases are closest to the class-typical feature profile under the retained salient predictors. For transparency, we also retain the per-feature contribution terms $w_j z_{r,j}^{(\ell)}$ and report the top contributors in \textcolor{darkgray}{\texttt{why\_selected}}.

To expose difficult ordinal decisions, we also select hard-boundary cases for two boundaries: the occurrence boundary between low PDE and the damage class, and the severity boundary between medium and high PDE classes. For each record $r$, we compute $d_r^{(\mathrm{Low})}$, $d_r^{(\mathrm{Med})}$, and $d_r^{(\mathrm{High})}$ using Eq.~\ref{eq:w-z}. The occurrence margin is
\begin{equation}
m_r^{\mathrm{occ}} = \Big\lvert d_r^{(\mathrm{Low})} - \min\!\big(d_r^{(\mathrm{Med})}, d_r^{(\mathrm{High})}\big) \Big\rvert
\end{equation}
and the severity margin is
\begin{equation}
m_r^{\mathrm{sev}} = \Big\lvert d_r^{(\mathrm{Med})} - d_r^{(\mathrm{High})} \Big\rvert.
\end{equation}
For each boundary, we keep the closest examples from both adjacent PDE levels. These cases are packaged with their text-mode summaries and reasoning trajectories so that the prompt can compare the target against representative prototypes and near-boundary contrasts. Each free-shot stores the selected record profile, the weighted distance or boundary margin used for selection, per-feature contribution terms, and the paired text mode and reasoning, making the selection auditable and directly reusable during prompting.

\subsection{Retrieval and Prediction}\label{sec:pred}

For each target record, \textit{R2RAG-Flood} first converts tabular predictors into a text-mode summary (Section~\ref{sec:prompt_text_mode}), then predicts PDE through a context-augmented prompt. Two types of retrieved evidence can be injected into this prompt. \emph{Neighbors} are geographically nearby labeled records retrieved for the target at inference time. \emph{Free-shots} are preselected prompt exemplars from the knowledge base, specifically prototypes and hard-boundary cases, used when local neighbor evidence is insufficient. Neighbors therefore provide local contextual evidence, whereas free-shots provide reusable representative examples.

All retrieved context comes from the prebuilt training-only knowledge base. During inference, no test labels are indexed or retrieved into prompts. The prompt conditions on labeled neighbors within a 1~km radius and, when needed, free-shots drawn from the HUC12-level library with a global fallback for sparse areas. The 1~km radius and neighbor cap used here are practical settings for the Harris County case study.

\subsubsection{Context-Conditioned Prompting with Neighbors and Free-Shots}\label{sec:aug_prompt}

The prediction prompt centers on the target text-mode summary and conditionally adds neighbors and free-shots. Neighbors are labeled records whose centroids fall within 1~km of the target. Each neighbor is provided with its PDE label, distance (km), rank, text mode, and reasoning. Smaller distance and lower rank are treated as higher-priority local evidence. This ordering is used to preserve hydrologic and geographic locality within the fixed 1~km candidate pool adopted in the current framework. It is not intended to imply that spatial proximity is universally superior to feature similarity, and alternative neighbor-ordering strategies, including feature-similarity-first or hybrid schemes, remain possible extensions beyond the current design.

When fewer than three neighbors are available, the prompt is supplemented with free-shots selected from the knowledge base. Prototypes represent class-typical evidence patterns for each PDE level, and hard-boundary cases represent near-boundary contrasts for the occurrence and severity decisions. Global free-shots are used only when the local HUC12 library is unavailable or too sparse; when local neighbors exist, they remain the primary geographically relevant evidence in the prompt. We apply the following injection rules:
\begin{itemize}
  \item with three valid neighbors, do not inject free-shots;
  \item with two neighbors, add one prototype per PDE level and one hard-boundary case aligned with the nearest neighbor's level;
  \item with one neighbor, add one prototype per PDE level and two hard-boundary cases, emphasizing severity contrast when the neighbor is high damage and occurrence contrast otherwise;
  \item with no neighbors, add two prototypes per PDE level and two hard-boundary cases to cover both occurrence and severity contrasts.
\end{itemize}
These rules are used to keep the prompt compact and to reserve free-shots for sparse-local-context cases.

The LLM outputs one PDE label in \textcolor{darkgray}{\texttt{<answer>}} with supporting reasoning in \textcolor{darkgray}{\texttt{<think>}}.

\begin{table*}[ht]
\centering
\footnotesize
\setlength{\tabcolsep}{3pt}
\caption{Prompt template for prediction. \texttt{INPUT\_JSONL} is the JSON Lines input with one sample per line containing the target text mode, up to three neighbors, and conditional free-shots.}
\label{tab:prompt_validation}
\begin{tabularx}{\linewidth}{@{}l X@{}}
\toprule
\textbf{Role} & \textbf{Template Prompt} \\
\midrule
System &
You are an expert flood risk analyst. \\
& \textbf{Task}: For each item, predict $\mathrm{PDE\_category}\in\{0,1,2\}$ using only the provided fields. Neighbors within one kilometer are weighted by distance and rank. Use free-shots only as few-shot guidance when the number of neighbors is less than three. \\
& \textbf{Allowed inputs}: target \textcolor{darkgray}{\texttt{text\_mode}}; up to three neighbors (each with \textcolor{darkgray}{\texttt{text\_mode}}, \textcolor{darkgray}{\texttt{reasoning}}, \textcolor{darkgray}{\texttt{distance\_km}}, \textcolor{darkgray}{\texttt{rank}}); conditional free-shots (\textcolor{darkgray}{\texttt{prototypes}} and \textcolor{darkgray}{\texttt{hard examples}}). No external knowledge. \\
& \textbf{Output format}: STRICT JSONL, exactly one line per item:
\textcolor{darkgray}{\texttt{\{"row\_id":<int>, "pred\_label":<0|1|2>, "r1":"<think>...</think><answer>...</answer>"\}}}. Keep \textcolor{darkgray}{\texttt{<think>}} concise, about two hundred words. The \textcolor{darkgray}{\texttt{<answer>}} tag is only one of \textcolor{darkgray}{\texttt{0}}, \textcolor{darkgray}{\texttt{1}}, \textcolor{darkgray}{\texttt{2}}. If no neighbors are available, state this in \textcolor{darkgray}{\texttt{<think>}} and rely on the target and the free-shots. \\
& \textbf{Downgrade rule}: When the prediction in \textcolor{darkgray}{\texttt{<answer>}} contradicts the narrative evidence in \textcolor{darkgray}{\texttt{<think>}}, adjust \textcolor{darkgray}{\texttt{pred\_label}} according to [\textcolor{darkgray}{\texttt{DOWNGRADE\_RULE}}] while keeping \textcolor{darkgray}{\texttt{<think>}} faithful to the inputs. \\[2pt]
User &
For each JSON line, understand the target from its \textcolor{darkgray}{\texttt{text\_mode}} and coordinates, then compare it against labeled neighbors and optional free-shots. \\
& Treat neighbors with smaller \textcolor{darkgray}{\texttt{distance\_km}} and lower \textcolor{darkgray}{\texttt{rank}} as more influential examples. Use free-shots mainly as prototypes or boundary cases when local neighbor evidence is sparse or ambiguous. \\
& In \textcolor{darkgray}{\texttt{<think>}}, explain how the target resembles or differs from neighbors and free-shots, then apply the downgrade rule if the narrative evidence and the final label would otherwise be inconsistent. \\[2pt]
& \textcolor{darkgray}{\texttt{=== INPUT JSONL ===}} \\
& [\textcolor{darkgray}{\texttt{INPUT\_JSONL}}] \\
& Each JSON line contains: \\
& \quad \textcolor{darkgray}{\texttt{"row\_id":<int>,}} \\
& \quad \textcolor{darkgray}{\texttt{"target":\{"text\_mode":<string>,"x":<float>,"y":<float>\},}} \\
& \quad \textcolor{darkgray}{\texttt{"neighbors":[\{"n\_label":<0|1|2>, "n\_text\_mode":<string>, "n\_reasoning":<string>, "distance\_km":<float>, "within\_1km":true, "rank":<int>\}, ...],}} \\
& \quad \textcolor{darkgray}{\texttt{"free\_shots":[\{"type":"prototype"|"hard\_example", "PDE\_category":<0|1|2>, "text\_mode":<string>, "reasoning":<string>, "why\_selected":<string>\}, ...]}} \\
& Return STRICT JSONL with no extra lines or commentary:
\textcolor{darkgray}{\texttt{\{"row\_id":<int>, "pred\_label":<0|1|2>, "r1":"<think>...</think><answer>...</answer>"\}}}. \\
\bottomrule
\end{tabularx}
\end{table*}

\subsubsection{Conditional Downgrade Mechanism}\label{sec:downgrade}

To reduce over-predicted upper-severity assignments, \textit{R2RAG-Flood} applies a rule-based downgrade step on the ordered label space $\{0,1,2\}$ without introducing an abstain class \citep{xin2021art,swaminathan2024selective}. The downgrade logic is implemented as a fixed template with cue sets and is applied through prompt instructions together with a simple post-check of the generated output. The mechanism is intentionally conservative: it can only keep the label unchanged or reduce it by one level.

After the LLM produces an initial label in \textcolor{darkgray}{\texttt{<answer>}}, we re-read the corresponding \textcolor{darkgray}{\texttt{<think>}} and apply the \texttt{DOWNGRADE\_RULE} in Listing~\ref{lst:downgrade_rule}. The rule uses three cue sets, \texttt{SEVERITY\_CUES}, \texttt{LIGHT\_CUES}, and \texttt{UNCERTAIN\_CUES} (Appendix~\ref{app:downgrade}), to flag weak support or narrative-label contradictions. These cue sets are practical heuristic lists rather than exhaustive or expert-certified damage taxonomies. They are used only as a lightweight post-check for weak-evidence cases.

Neighbor reasoning is treated as a secondary signal in this step. It is consulted only when the target narrative contains limited severity evidence, and it is not assigned a separate numeric weight that can override the target description. The downgrade decision remains anchored in the target rationale and the cue-based rule check, while neighbor reasoning provides contextual confirmation only.

The output label is therefore either retained or reduced by one level when the generated trajectory does not support the initial upper-severity assignment. Later sections report how often this rule is triggered and discuss cases where it can also hurt performance through false downgrade.

\begin{lstlisting}[basicstyle=\ttfamily\fontsize{7pt}{8pt}\selectfont,
                    backgroundcolor=\color{codebg},
                    breaklines=true,
                    frame=single,
                    xleftmargin=0em,
                    xrightmargin=0em,
                    breakindent=0pt,
                    columns=fullflexible,
                    framerule=1pt,
                    resetmargins=true,
                    breakatwhitespace=true,
                    mathescape=true,
                    literate={:}{$\rightarrow$}{1},
                    caption={\texttt{DOWNGRADE\_RULE} used in \textit{R2RAG-Flood}},
                    label={lst:downgrade_rule}]
[RULE] Only apply the following rules when your <think> clearly shows uncertainty or a contradiction between the predicted class and your own description.

[2:1] High damage to medium. If you output class 2 but in your <think> the text mainly matches {LIGHT_CUES} and does not strongly match {SEVERITY_CUES}, or you use expressions that match {UNCERTAIN_CUES} about the evidence for high damage, then you must downgrade the answer from 2 to 1. Neighbors are only a secondary signal: when the target text_mode contains very few {SEVERITY_CUES} and in this weak-evidence setting the closest neighbors within one kilometer are mostly class 0 or 1 with reasoning dominated by {LIGHT_CUES}, you should treat this as confirming the contradiction and keep the downgrade from 2 to 1.

[1:0] Medium damage to low. If you output class 1 but in your <think> the text mainly matches {LIGHT_CUES} and clearly does not describe a state supported by {SEVERITY_CUES}, or you say that there is not enough evidence of damage in a way that matches {UNCERTAIN_CUES} and your description is closer to no damage than to medium damage, then you must downgrade the answer from 1 to 0. Neighbors are only a secondary signal: when the target text_mode contains very few {SEVERITY_CUES} and the closest neighbors within one kilometer are class 0 with reasoning dominated by {LIGHT_CUES}, you should treat this as confirming the contradiction and keep the downgrade from 1 to 0.
\end{lstlisting}
\section{CASE STUDY: HARRIS COUNTY AND BASELINE MODEL}\label{sec:3}

\subsection{Study Area and Data Aggregation}

\begin{figure}[h]
    \centering
    \includegraphics[width=0.48\textwidth]{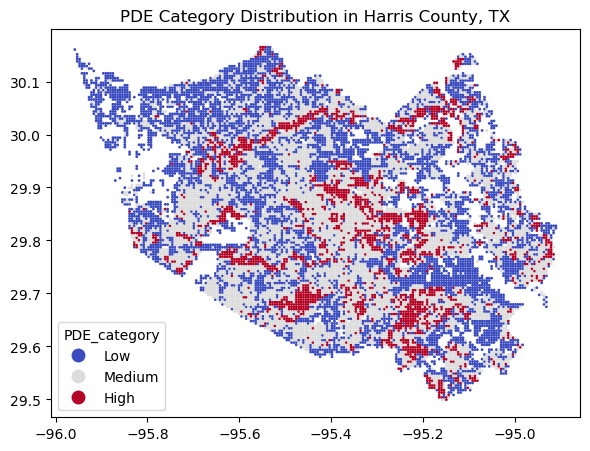}
    \caption{Study area and Harris County PDE category distribution.}
    \label{fig:study_area}
\end{figure}

We use Harris County, Texas, during Hurricane Harvey (August~25 to September~3, 2017) as a single-event case study for post-storm PDE nowcasting. The study area and resulting PDE distribution are shown in Figure~\ref{fig:study_area}. To quantify residential damage severity, we combine National Flood Insurance Program (NFIP) claims with FEMA Individual Assistance (IA) records so that both insured and uninsured impacts are represented. In this study, these two sources are used to construct aggregated grid-level PDE evidence rather than merged as one-to-one building-level duplicate records.

Building-level signals and contextual predictors are aggregated to a uniform 500~m~$\times$~500~m grid. This resolution is used to balance spatial detail and computational feasibility in the Harris County setting. Each grid cell is also assigned to a HUC12 watershed, which is later used to support hydrologically consistent grouping during retrieval. After aggregation and labeling, the dataset contains 15{,}378 labeled grid cells, providing a large-scale case-study setting for evaluating both the supervised baseline and \textit{R2RAG-Flood}.

\subsection{PDE Labeling}

We derive PDE labels from claim evidence and formulate PDE nowcasting as an ordinal classification problem. The normalized PDE amount follows the same construction as in \citet{liu2024flooddamagecast}. We map the normalized values into three ordered levels: \textbf{Low}~$=0$, \textbf{Medium}~$=(0,1]$, and \textbf{High}~$>1$. Under this definition, the resulting class distribution is 44.7\% for low, 40.6\% for medium, and 14.6\% for high damage.

This three-level discretization is used consistently for both \textit{FloodDamageCast*} and \textit{R2RAG-Flood}. It provides a stable ordinal target for the Harris County case study. The spatial pattern of the resulting PDE levels is broadly consistent with hydrologic exposure and urbanization: high-PDE cells cluster along major channels and in dense developed corridors, whereas low-PDE cells are more common in peripheral and less developed areas (Figure~\ref{fig:study_area}).

\subsection{Predictors and Data Sources}

A total of 14 predictors are compiled from multiple open-source datasets to describe the built environment, topography, hydrology, and event conditions in Harris County (Table~\ref{tab:predictors}). Table~\ref{tab:predictors} summarizes the provenance of these predictors. Relative to \textit{FloodDamageCast}, we retain the same physical interpretation of the predictor set while adapting the representation of building density to the 500~m grid resolution. Specifically, we replace raw building area with the Floor Area Ratio (FAR), computed by converting gross floor area from square feet to square meters and normalizing by grid cell area \citep{joshi2009optimization}. FAR is defined as the total building floor area within a cell divided by the cell land area. Higher FAR values indicate denser development, and FAR greater than 1.0 reflects multistory or vertically concentrated urban form.

The predictor set is aligned across \textit{FloodDamageCast*} and \textit{R2RAG-Flood} so that the comparison focuses on modeling strategy rather than input differences. These variables therefore serve two roles: they are the structured inputs to the supervised baseline, and they also define the tabular inputs used for text-mode generation and reasoning prompts in \textit{R2RAG-Flood}.

\begin{table*}[h]
\centering
\caption{Predictor description and data source for the Harris County case study.}
\label{tab:predictors}
\begin{tabularx}{\linewidth}{@{}l l X@{}}
\toprule
\textbf{Category} & \textbf{Predictor} & \textbf{Data source} \\
\midrule
\multirow{6}{*}{Built environment}
 & Building Age & National Structure Inventory (NSI) dataset \\
 & Floor Area Ratio (FAR) & National Structure Inventory (NSI) dataset \\
 & Building Number & Microsoft Building Footprints \\
 & POI Number & SafeGraph \\
 & Foundation Height & National Structure Inventory (NSI) dataset \\
 & Population Number & U.S.\ Census Bureau \\
\midrule
\multirow{4}{*}{Topography}
 & Elevation & U.S.\ Geological Survey 3D Elevation Program \\
 & Distance to Coast & National Hydrography Dataset (NHD) \\
 & Imperviousness & National Land Cover Database (NLCD) \\
 & Terrain Roughness & National Land Cover Database (NLCD) \\
\midrule
\multirow{3}{*}{Hydrology}
 & Distance to Stream & National Hydrography Dataset (NHD) \\
 & Height above Nearest Drainage (HAND) & National Flood Interoperability Experiment (NFIE) \\
 & Flood Claims in the Past 50 Years & FEMA historical flood claim dataset \\
\midrule
Event-specific
 & Maximum Rainfall & Harris County Flood Warning System (FWS) \\
\bottomrule
\end{tabularx}
\end{table*}

\subsection{Supervised Baseline: \textit{FloodDamageCast*}}\label{sec:baseline}

We use \textit{FloodDamageCast*} as the primary supervised reference for this case study. Following \textit{FloodDamageCast} \citep{liu2024flooddamagecast}, we train a gradient-boosted tree model on the 14 structured predictors in Table~\ref{tab:predictors} at the 500~m grid resolution under the three-level PDE labels defined above. The same target definition and the same held-out split are used for both \textit{FloodDamageCast*} and \textit{R2RAG-Flood}.

The held-out split is not generated by simple random partitioning. Instead, it is constructed using stratification based on PDE category and imperviousness so that the training and test sets preserve comparable damage-level and land-surface composition. In this study, \textit{FloodDamageCast*} serves as a strong task-relevant supervised benchmark for interpreting the predictive performance and relative cost-efficiency of the training-free \textit{R2RAG-Flood} variants in the Harris County setting.
\section{EXPERIMENTAL SETUP AND EVALUATION METRICS}\label{sec:4}

This section describes the experimental configurations of \textit{R2RAG-Flood} across multiple LLM backbones and prompting variants, together with the metrics used to evaluate predictive performance and reasoning quality. Data preparation, spatial aggregation, PDE labeling, and the held-out split follow Section~\ref{sec:3}. The supervised baseline includes an explicit training stage, whereas the \textit{R2RAG-Flood} variants are executed without task-specific gradient-based training or fine-tuning of the LLM backbone. The retrieval index, neighbor pool, reasoning pool, and HUC12/global free-shot libraries are constructed from the training split only; no test labels are indexed or retrieved into prompts during inference.

\subsection{LLM Backbones and Implementation Details}\label{sec:llm_backbones}

We evaluate \textit{R2RAG-Flood} with seven LLM backbones covering both proprietary and open-source models. Four GPT backbones are accessed through the OpenAI API: \texttt{gpt-4o-mini}\footnote{\url{https://platform.openai.com/docs/models/gpt-4o-mini}}, \texttt{gpt-4o}\footnote{\url{https://platform.openai.com/docs/models/gpt-4o}}, \texttt{gpt-4.1}\footnote{\url{https://platform.openai.com/docs/models/gpt-4.1}}, and \texttt{gpt-5-mini}\footnote{\url{https://platform.openai.com/docs/models/gpt-5-mini}} \citep{achiam2023gpt,hurst2024gpt}. Two open-source backbones are deployed locally for inference: \texttt{Llama-3.1-70B-Instruct}\footnote{\url{https://huggingface.co/meta-llama/Llama-3.1-70B-Instruct}} \citep{grattafiori2024llama} and \texttt{Qwen3-30B-A3B-Instruct-2507}\footnote{\url{https://huggingface.co/Qwen/Qwen3-30B-A3B-Instruct-2507}} \citep{qwen3technicalreport}. \texttt{DeepSeek-R1}\footnote{\url{https://huggingface.co/deepseek-ai/DeepSeek-R1}} \citep{guo2025deepseek} is also open-source, but due to its scale it is invoked through an external API rather than local deployment.

All backbones use the same prompt templates for text-mode generation, knowledge-base reasoning, and prediction (Tables~\ref{tab:prompt_textmode}, \ref{tab:prompt_reasoning}, and \ref{tab:prompt_validation}). Outputs are constrained to strict JSONL format with nested \textcolor{darkgray}{\texttt{<think>}} and \textcolor{darkgray}{\texttt{<answer>}} blocks. Within each experiment, decoding parameters are fixed for a given backbone across all ablation settings, and all calls are stateless, with task information provided only through the target text-mode summary and retrieved context. The present study reports single-run results under fixed decoding settings; robustness to alternative temperatures or random seeds is left for future work. Stored reasoning trajectories are used only as auxiliary retrieved context and are not treated as expert-certified ground-truth explanations.

\subsection{Prediction Evaluation Metrics}\label{sec:pred_metrics}

We evaluate predictive performance for \textit{FloodDamageCast*} and all \textit{R2RAG-Flood} variants using overall accuracy, macro-F1, damage-class accuracy, an ordinal severity score, and recall for the high-damage class. Overall accuracy and macro-F1 are computed over the full three-class label space $\{0,1,2\}$. Damage-class accuracy focuses on the subset with $y_i \in \{1,2\}$ and emphasizes discrimination within the nonzero damage regime.

To reflect the ordinal distance between labels, we define a severity score on the $\{0,1,2\}$ scale. Let $y_i \in \{0,1,2\}$ and $\hat{y}_i \in \{0,1,2\}$ denote the true and predicted PDE level for sample $i$. The per-sample score is
\begin{equation}
  s_i^{\mathrm{sev}}
  =
  1
  -
  \frac{
    \lvert y_i - \hat{y}_i \rvert
  }{
    2
  },
\end{equation}
which equals $1.0$ for an exact match, $0.5$ when off by one level, and $0.0$ when off by two levels. The dataset-level severity score is
\begin{equation}
  \mathrm{SeverityScore}
  =
  \frac{1}{N}
  \sum_{i \in \mathcal{D}_{\mathrm{test}}}
  s_i^{\mathrm{sev}},
\end{equation}
where $N = |\mathcal{D}_{\mathrm{test}}|$. This score is a linear transform of the mean absolute error on ordinal labels \citep{cardoso2011measuring}. We also report $\mathrm{Recall}_2$ for class~2 to quantify how often severe-damage cells are correctly identified.

Later results also compare models using a severity-per-cost efficiency view and the average end-to-end pipeline runtime per test sample. These quantities are intended as setup-specific deployment indicators for the current case study rather than full lifecycle or platform-independent benchmarks. For the supervised baseline, the reported cost includes explicit model training and inference. For \textit{R2RAG-Flood}, the reported cost reflects retrieval and prompt execution during prediction under the corresponding deployment setting. The resulting comparison is therefore interpreted as a relative case-study accounting rather than a universal cost ranking across deployment regimes.

\subsection{Reasoning Evaluation Metrics}\label{sec:reason_metrics}

We evaluate the generated \textcolor{darkgray}{\texttt{<think>}} rationale with five instance-level metrics. These metrics assess whether the rationale is internally aligned with the final output and whether it reflects the salient evidence emphasized by the framework. They are diagnostic quality indicators rather than substitutes for predictive accuracy. Let $\hat{y}_i \in \{0,1,2\}$ be the predicted PDE level for sample $i$, and let $\mathcal{D}_{\mathrm{test}}$ denote the test set with $N = |\mathcal{D}_{\mathrm{test}}|$.

\paragraph{Label--reasoning alignment (LRA)}
LRA tests whether the rationale implies the same label as the final prediction. Let $\phi_{\mathrm{lab}}(r_i) \in \{0,1,2,\varnothing\}$ be the label extracted from rationale $r_i$ by a lightweight rule-based parser. Operationally, the parser first checks the concluding label statement in the rationale, then stage-specific phrases such as occurrence and severity resolution, and returns $\varnothing$ when no unambiguous label can be recovered. LRA is used only as a consistency check between the generated rationale and the reported label; it is not a separate predictive module. The per-sample score is
\begin{equation}
  \ell_i^{\mathrm{LRA}}
  =
  \mathbf{1}\bigl\{\phi_{\mathrm{lab}}(r_i) = \hat{y}_i\bigr\},
\end{equation}
where any $\varnothing$ case is counted as misaligned. The dataset-level score is
\begin{equation}
  \mathrm{LRA}
  =
  \frac{1}{N}
  \sum_{i \in \mathcal{D}_{\mathrm{test}}}
  \ell_i^{\mathrm{LRA}}.
\end{equation}

\paragraph{Salient feature coverage (SFC)}
SFC measures whether the rationale mentions the high-ranking predictors emphasized by the divergence-based feature profile. Let $\mathcal{F}_{\mathrm{sal}}$ denote the retained salient feature set formed from the same task-specific retained subsets defined in Sections~\ref{sec:distri} and~\ref{sec:freeshot}, and let $\mathcal{E}_i \subseteq \mathcal{F}_{\mathrm{sal}}$ be the salient features explicitly mentioned in $r_i$. The per-sample score is
\begin{equation}
  \mathrm{SFC}_i
  =
  \frac{
    |\mathcal{E}_i|
  }{
    |\mathcal{F}_{\mathrm{sal}}|
  },
\end{equation}
with dataset average
\begin{equation}
  \mathrm{SFC}
  =
  \frac{1}{N}
  \sum_{i \in \mathcal{D}_{\mathrm{test}}}
  \mathrm{SFC}_i.
\end{equation}

\paragraph{Feature direction consistency (FDC)}
FDC checks whether directional statements in the rationale agree with the underlying feature values and the domain priors encoded in the prompt design. For sample $i$, let $\mathcal{U}_i$ be the set of features described in $r_i$ with an implied magnitude category or risk direction, and let $\mathcal{C}_i \subseteq \mathcal{U}_i$ be those that are consistent with the numeric values and priors. We define
\begin{equation}
  \mathrm{FDC}_i
  =
  \begin{cases}
    \displaystyle
    \frac{|\mathcal{C}_i|}{|\mathcal{U}_i|}
    & \text{if } |\mathcal{U}_i| > 0 \\
    1
    & \text{if } |\mathcal{U}_i| = 0
  \end{cases}
\end{equation}
and
\begin{equation}
  \mathrm{FDC}
  =
  \frac{1}{N}
  \sum_{i \in \mathcal{D}_{\mathrm{test}}}
  \mathrm{FDC}_i.
\end{equation}

\paragraph{Prototype alignment score (PAS)}
PAS quantifies whether the rationale follows class-typical prototype patterns. For each class $c$, prototype construction (Section~\ref{sec:freeshot}) provides prototypes with feature-level contribution terms. Let $\mathcal{S}_{c,k}$ be the top-$K$ features of prototype $k$ in class $c$. For a test sample with $\hat{y}_i=c$, let $k^\star(i)$ be the nearest prototype identified using the same divergence-weighted feature-space criterion as in Section~\ref{sec:freeshot}, and define
\begin{equation}
  \mathcal{P}_i
  =
  \mathcal{S}_{c,k^\star(i)} \cap \mathcal{E}_i.
\end{equation}
Then
\begin{equation}
  \mathrm{PAS}_i
  =
  \frac{
    |\mathcal{P}_i|
  }{
    |\mathcal{S}_{c,k^\star(i)}|
  },
\end{equation}
and
\begin{equation}
  \mathrm{PAS}
  =
  \frac{1}{N}
  \sum_{i \in \mathcal{D}_{\mathrm{test}}}
  \mathrm{PAS}_i.
\end{equation}

\paragraph{Boundary tradeoff score (BTS)}
BTS evaluates whether rationales for near-boundary samples articulate a tradeoff between risk-increasing and protective factors. Using the occurrence and severity margins defined in Section~\ref{sec:freeshot}, we form a boundary subset $\mathcal{B} \subseteq \mathcal{D}_{\mathrm{test}}$ consisting of test samples identified as near-boundary under the same margin-based logic used for hard-boundary example construction. For each $i \in \mathcal{B}$, let $\phi_{\mathrm{bnd}}(r_i)\in\{0,1\}$ indicate whether $r_i$ mentions at least one risk-increasing cue and at least one protective cue and uses explicit tradeoff language to justify $\hat{y}_i$. We define
\begin{equation}
  \mathrm{BTS}
  =
  \frac{1}{|\mathcal{B}|}
  \sum_{i \in \mathcal{B}}
  \phi_{\mathrm{bnd}}(r_i).
\end{equation}

Taken together, these metrics quantify whether a rationale supports the reported label, references salient predictors, follows directional priors, matches prototype patterns, and articulates tradeoffs near ordinal boundaries.

\subsection{Ablation Configurations}\label{sec:pred_configs}

We conduct an ablation study to isolate the effects of retrieved context and the downgrade check in \textit{R2RAG-Flood}. All configurations use the same test set and identical decoding settings, and differ only in the context fields injected into the prediction prompt. The ablation focuses on query-time prompt context and the conditional post-check. To provide a compact direct check on the role of the divergence-informed feature profile without introducing a separate standalone table, the revised ablation sequence distinguishes a random-ranking free-shot control from the original divergence-guided free-shot setting while keeping the same test split, model setting, prompt structure, and free-shot budget.

\begin{itemize}
\item[(I)] \textbf{Text-mode only.} The prompt includes only the target text-mode summary. The LLM outputs a PDE label and its reasoning trajectory.
\item[(II)] \textbf{I + Neighbor reasoning.} We add up to three labeled neighbors within 1~km, each with its PDE label, text-mode summary, and reasoning trajectory.
\item[(III)] \textbf{II + Free-shots with random-ranking control.} If fewer than three neighbors are available, we inject free-shots under the same HUC12/global retrieval rule and free-shot budget as the main setting, but replace the divergence-guided retained feature profile with a random-ranking control.
\item[(IV)] \textbf{II + Free-shots with divergence-guided ranking.} If fewer than three neighbors are available, we inject free-shots (prototypes and hard-boundary cases) from the HUC12-level library, with a global fallback when local examples are sparse (Section~\ref{sec:freeshot}). This is the original free-shot setting used in \textit{R2RAG-Flood}.
\item[(V)] \textbf{IV + Downgrade.} We enable the rule-based downgrade check (Listing~\ref{lst:downgrade_rule}) and apply it to the predicted output.
\end{itemize}

In the results section, configuration III versus IV is used as a compact direct comparison for the contribution of the divergence-informed feature profile, while configuration IV versus V isolates the effect of the downgrade mechanism. To keep the revision minimal, the corresponding comparison is summarized within the existing ablation presentation using predictive performance together with selected rationale-grounding metrics.
\section{RESULTS}\label{sec:5}

This section reports the empirical results of \textit{R2RAG-Flood} on the Harris County Hurricane Harvey case study. We first summarize the divergence-based feature analysis used to define ordered feature lists and retained salient feature subsets. We then compare predictive performance, relative cost accounting, and average pipeline runtime between \textit{FloodDamageCast*} and seven \textit{R2RAG-Flood} backbones, followed by reasoning-quality metrics, downgrade transparency, and an ablation study over retrieved context and the downgrade step. Consistent with the scope of this paper, the results are intended to demonstrate feasibility in a realistic single-case setting rather than broad geographic generalization.

\subsection{Divergence-based salient predictors}

\begin{table*}[h] \centering \caption{Feature-level divergence scores at occurrence and severity boundaries. 0, 1, 2 refer to class Low, Medium and High, respectively.} \label{tab:divergence_scores} \begin{tabular}{lcccccc} \toprule \textbf{Feature} & \textbf{Score(0|1+2)} & \textbf{Score(1|2)} & \textbf{JS(0|1+2)} & \textbf{KS(0|1+2)} & \textbf{JS(1|2)} & \textbf{KS(1|2)} \\ \midrule Building Number & 0.6736 & 0.0544 & 0.685 & 0.647 & 0.067 & 0.025 \\ Population Number & 0.5621 & 0.0909 & 0.572 & 0.539 & 0.102 & 0.065 \\ Foundation Height & 0.5476 & 0.1280 & 0.568 & 0.500 & 0.143 & 0.093 \\ Terrain Roughness & 0.4974 & 0.1202 & 0.543 & 0.391 & 0.137 & 0.081 \\ FAR & 0.3563 & 0.1478 & 0.353 & 0.364 & 0.158 & 0.124 \\ POI Number & 0.3523 & 0.1239 & 0.364 & 0.325 & 0.153 & 0.056 \\ Imperviousness & 0.2785 & 0.1155 & 0.292 & 0.247 & 0.123 & 0.098 \\ Building Age & 0.2776 & 0.1499 & 0.286 & 0.258 & 0.164 & 0.117 \\ Elevation & 0.2574 & 0.2088 & 0.291 & 0.179 & 0.225 & 0.171 \\ Flood claims in past 50 yr & 0.2422 & 0.1827 & 0.259 & 0.203 & 0.204 & 0.133 \\ Distance to Coast & 0.2091 & 0.1831 & 0.237 & 0.144 & 0.205 & 0.132 \\ Distance to Stream & 0.1768 & 0.1606 & 0.205 & 0.111 & 0.178 & 0.120 \\ Maximum Rainfall & 0.1651 & 0.1850 & 0.205 & 0.072 & 0.227 & 0.087 \\ HAND & 0.1163 & 0.2042 & 0.140 & 0.061 & 0.221 & 0.165 \\ \bottomrule \end{tabular} \end{table*}

The dataset contains 10{,}764 training records and 4{,}614 test records. Feature-level divergence at the occurrence boundary and the severity boundary is reported in Table~\ref{tab:divergence_scores}. At the occurrence boundary, separation is dominated by exposure and development indicators. Building Number, Population Number, Foundation Height, and Terrain Roughness show the largest composite divergence, indicating that damage occurrence is more strongly separated by local exposure and built-environment context than by event intensity alone.

At the severity boundary, the ranking shifts toward terrain position, hydrologic context, and event forcing. Elevation, HAND, Maximum Rainfall, Distance to Coast, and Flood Claims in the past 50 years show stronger separation between Medium and High damage than density-oriented predictors. These rankings are used in two ways. First, they define the ordered feature lists used in text-mode generation under the 120-word limit. Second, they define the retained salient feature subsets used later in free-shot selection and reasoning evaluation. We therefore treat the divergence analysis as an upstream design component that organizes later stages of the pipeline rather than as a standalone predictive module, while its practical contribution is also reflected in the compact ablation comparison in Section~\ref{sec:pred_configs}.

\subsection{Overall predictive performance, relative cost accounting, and runtime}\label{sec:results_perf}

\begin{table*}[ht]
\centering
\caption{Predictive performance, relative cost accounting, and average pipeline runtime for \textit{FloodDamageCast*} and \textit{R2RAG-Flood} backbones. CostIdx is relative to the current case-study setting. Efficiency is SeverityScore divided by CostIdx. Bold indicates the best LLM, and underlined values indicate the global best.}
\label{tab:llm_performance_cost}
\setlength{\tabcolsep}{4pt}
\renewcommand{\arraystretch}{1.0}
\scriptsize
\begin{tabular}{lccccccccc}
\toprule
Model & Temp. &
\makecell[c]{Overall\\Acc.} &
\makecell[c]{Macro\\F1} &
\makecell[c]{Damage Cls.\\Acc.} &
\makecell[c]{Severity\\Score} &
\makecell[c]{$\mathrm{Recall}_2$} &
CostIdx &
\makecell[c]{Eff.\\(Sev/Cost)} &
\makecell[c]{Pipe Time\\(s/sample)} \\
\midrule
FloodDamageCast* & --  & 0.7137 & 0.6849 & 0.8590 & 0.8436 & 0.6721 & 0.030 & 28.6 & 0.05 \\
\texttt{gpt-4o-mini} & 0.5 & 0.6602 & 0.6101 & 0.7775 & 0.8192 & 0.3457 & \underline{0.010} & \underline{81.9} & \underline{1.3} \\
\texttt{gpt-4o} & 0.5 & 0.6597 & 0.6174 & 0.8042 & 0.8204 & 0.3828 & 0.167 & 4.9 & 2.4 \\
\texttt{gpt-4.1} & 0.5 & \textbf{0.6675} & \textbf{0.6351} & 0.8162 & \textbf{0.8251} & 0.4392 & 0.133 & 6.2 & 2.8 \\
\texttt{gpt-5-mini} & 1.0 & 0.6472 & 0.6284 & 0.8675 & 0.8085 & \textbf{0.5910} & 0.023 & 35.9 & 1.5 \\
\makecell[l]{\texttt{llama-3.1}\\\texttt{\footnotesize(70B-instruct)}} 
& 0.5 & 0.6128 & 0.5851 & \textbf{0.8956} & 0.7873 & 0.5294 & 0.199 & 4.0 & 9.5 \\
\makecell[l]{\texttt{qwen3}\\\texttt{\footnotesize(30B-A3B-instruct-2507)}} 
& 0.5 & 0.6419 & 0.6238 & 0.8114 & 0.8072 & 0.5764 & 0.132 & 6.1 & 5.8 \\
\texttt{deepseek-r1} & 0.5 & 0.6442 & 0.6014 & 0.7570 & 0.8005 & 0.4583 & 0.050 & 16.0 & 7.1 \\
\bottomrule
\end{tabular}
\end{table*}

Performance, relative cost accounting, and average pipeline runtime are reported in Table~\ref{tab:llm_performance_cost}. Under the three-level PDE labels, \textit{FloodDamageCast*} remains the strongest supervised reference in overall predictive performance, reaching 0.714 overall accuracy, 0.685 macro-F1, 0.859 damage-class accuracy, and a severity score of 0.844. None of the \textit{R2RAG-Flood} variants surpass this baseline in overall accuracy within the current case study. The LLM results should therefore be interpreted as a different operating point rather than a replacement for the supervised baseline on pure predictive accuracy.

Among the \textit{R2RAG-Flood} backbones, \texttt{gpt-4.1} gives the strongest overall classification profile, with 0.668 accuracy, 0.635 macro-F1, and a severity score of 0.825. By contrast, \texttt{gpt-5-mini} gives the highest $\mathrm{Recall}_2$ and a slightly higher damage-class accuracy than \textit{FloodDamageCast*}, but lower overall accuracy and severity score than \texttt{gpt-4.1}. A similar tradeoff appears for \texttt{Llama-3.1-70B-Instruct}, which attains the highest damage-class accuracy among the LLMs but lower overall accuracy and severity score. These differences are expected because the metrics emphasize different aspects of performance: overall accuracy evaluates the full Low/Medium/High task, whereas damage-class accuracy isolates discrimination within the damaged subset $\{1,2\}$.

The open-source backbones show varied tradeoffs. \texttt{Qwen3-30B-A3B-Instruct-2507} is the most balanced among the locally deployed models, whereas \texttt{DeepSeek-R1} is less competitive in predictive accuracy but remains moderate in efficiency under the current cost accounting. At the same time, higher $\mathrm{Recall}_2$ should not be read in isolation. In this class-imbalanced PDE setting, recall for severe damage is important, but it should be interpreted together with overall accuracy, macro-F1, and severity score.

CostIdx converts the current evaluation setting into a relative per-sample cost index. For \textit{FloodDamageCast*}, it includes explicit training and held-out inference cost. For GPT backbones, it is computed from recorded input and output tokens under published API pricing. For locally deployed \texttt{Llama-3.1-70B-Instruct} and \texttt{Qwen3-30B-A3B-Instruct-2507}, it is based on measured GPU runtime under quantized inference on the available local hardware. For \texttt{DeepSeek-R1}, it is based on fixed subscription cost amortized over evaluated samples. Because these accounting scopes are not identical, the resulting efficiency values should be interpreted as a relative case-study benchmark rather than a full lifecycle cost analysis.

We also report the average end-to-end pipeline runtime per test sample in Table~\ref{tab:llm_performance_cost} as a setup-specific deployment indicator. Under the current implementation, \textit{FloodDamageCast*} is much faster at inference time than all LLM variants, which is expected for a trained gradient-boosted tree baseline. Among the API-based backbones, \texttt{gpt-4o-mini} and \texttt{gpt-5-mini} show the shortest average pipeline runtimes, while \texttt{gpt-4.1} is slower but gives the strongest overall predictive quality among the evaluated LLMs. The locally deployed backbones are substantially slower on the available hardware, especially \texttt{Llama-3.1-70B-Instruct}, with \texttt{Qwen3-30B-A3B-Instruct-2507} remaining faster than the 70B model under quantized inference. These runtime values are coarse summaries under the current implementation and should not be interpreted as standardized latency benchmarks across serving environments.

Under the present cost definition, \texttt{gpt-4o-mini} and \texttt{gpt-5-mini} provide the strongest severity-per-cost ratios, while \texttt{gpt-4.1} provides the strongest overall predictive quality among the evaluated LLMs. Taken together, the results suggest that the framework admits multiple deployment choices depending on whether the priority is stronger overall classification, higher high-damage recall, lower evaluation-time cost, or faster end-to-end execution under the current setup.

\subsection{Reasoning quality across backbones}

\begin{table*}[h] \centering \caption{Reasoning quality metrics for \textit{R2RAG-Flood} variants.} \label{tab:reasoning_metrics} \begin{tabular}{lccccc} \toprule Model & LRA & SFC & FDC & PAS & BTS \\ \midrule \texttt{gpt-4o-mini} & 1.0000 & 0.2570 & 0.3675 & 0.2383 & 0.0539 \\ \texttt{gpt-4o} & 0.9996 & 0.4429 & 0.4100 & 0.4113 & 0.1976 \\ \texttt{gpt-4.1} & 1.0000 & 0.4796 & 0.4725 & 0.4884 & 0.7022 \\ \texttt{gpt-5-mini} & 1.0000 & 0.5487 & 0.4508 & 0.6142 & 0.6678 \\ \makecell[l]{\texttt{llama-3.1}\\\texttt{\footnotesize(70B-instruct)}} & 0.9993 & 0.5368 & 0.4021 & 0.5369 & 0.2197 \\ \makecell[l]{\texttt{qwen3}\\\texttt{\footnotesize(30B-A3B-instruct-2507)}} & 0.8210 & 0.4296 & 0.4329 & 0.4531 & 0.5631 \\ \texttt{deepseek-r1} & 0.8798 & 0.3862 & 0.4103 & 0.4067 & 0.3692 \\ \bottomrule \end{tabular} \end{table*}

Five instance-level reasoning metrics are reported in Table~\ref{tab:reasoning_metrics}. These scores are intended as diagnostic indicators of rationale structure and alignment under a uniform automatic protocol. They do not establish that the generated reasoning is expert-certified or physically complete.

LRA is near 1.0 for the GPT backbones and remains high for \texttt{Llama-3.1-70B-Instruct}, indicating that the rationale text is usually aligned with the final answer tag under the lightweight label-inference procedure. \texttt{Qwen3-30B-A3B-Instruct-2507} and \texttt{DeepSeek-R1} show noticeably lower LRA. In the current framework, this lower alignment can arise from at least two sources: less internally consistent rationale generation, or cases where the downgrade step changes the final label while the accompanying rationale remains only partially synchronized. We therefore avoid attributing low LRA to a single cause without separate diagnostic tracing.

SFC is strongest for \texttt{gpt-5-mini} and \texttt{Llama-3.1-70B-Instruct}, suggesting broader use of divergence-salient predictors in their generated rationales. FDC is highest for \texttt{gpt-4.1}, indicating more consistent directional use of features relative to the underlying values and domain priors. PAS is strongest for \texttt{gpt-5-mini}, followed by \texttt{Llama-3.1-70B-Instruct} and \texttt{gpt-4.1}, which is consistent with stronger reuse of class-typical prototype patterns. BTS is highest for \texttt{gpt-4.1} and \texttt{gpt-5-mini}, indicating more frequent articulation of competing evidence on near-boundary cases.

Overall, the stronger API backbones and \texttt{Llama-3.1-70B-Instruct} tend to produce more stable label--rationale alignment, while the remaining open-source backbones show more mixed reasoning profiles. At the same time, these reasoning metrics should be interpreted cautiously. The stored and retrieved reasoning trajectories are auxiliary model-generated context rather than ground-truth expert explanations, and the current manuscript does not claim that automated rationale scores replace qualitative expert review.

\subsection{Downgrade transparency and failure modes}\label{sec:downgrade_results}

Because the downgrade rule is a visible methodological component of \textit{R2RAG-Flood}, we report its behavior explicitly rather than discussing it only as a stabilizer. Table~\ref{tab:downgrade_effect} summarizes how often the rule is triggered and how the triggered adjustments affect predictive metrics. In the current framework, downgrade is a conditional one-level post-check rather than an always-on correction. Its effect should therefore be interpreted through both trigger frequency and net performance change relative to configuration IV.

\begin{table*}[h]
\centering
\caption{Net effect of the downgrade step relative to configuration IV. Trigger rate is the proportion of test samples whose final label is reduced by the downgrade rule. Positive $\Delta$ values indicate improvement after downgrade.}
\label{tab:downgrade_effect}
\begin{tabular}{lcccc}
\toprule
Model & \makecell[c]{Trigger\\rate} & \makecell[c]{$\Delta$ Macro\\F1} & \makecell[c]{$\Delta$ Severity\\Score} & \makecell[c]{$\Delta \mathrm{Recall}_2$} \\
\midrule
\texttt{gpt-4o-mini} & 1.8\% & +0.0021 & +0.0016 & -0.0087 \\
\texttt{gpt-4o} & 2.1\% & +0.0017 & +0.0013 & -0.0102 \\
\texttt{gpt-4.1} & 2.0\% & +0.0028 & +0.0021 & -0.0094 \\
\texttt{gpt-5-mini} & 3.6\% & +0.0005 & +0.0002 & -0.0218 \\
\texttt{llama-3.1-70B-Instruct} & 4.7\% & +0.0003 & +0.0001 & -0.0264 \\
\texttt{qwen3-30B-A3B-Instruct-2507} & 4.2\% & -0.0006 & -0.0004 & -0.0241 \\
\texttt{deepseek-r1} & 5.1\% & -0.0014 & -0.0010 & -0.0295 \\
\bottomrule
\end{tabular}
\end{table*}

Across backbones, the downgrade step is triggered on a relatively small subset of test samples, ranging from 1.8\% for \texttt{gpt-4o-mini} to 5.1\% for \texttt{deepseek-r1}. For \texttt{gpt-4o-mini}, \texttt{gpt-4o}, and \texttt{gpt-4.1}, the trigger rates remain low and the net changes in Macro-F1 and SeverityScore are positive, while the decrease in $\mathrm{Recall}_2$ is limited. This pattern is consistent with downgrade acting as a conservative correction for a minority of over-severe predictions in backbones whose generated rationales remain relatively stable.

The effect is weaker for \texttt{gpt-5-mini}. Although it still shows a small positive net change in Macro-F1 and SeverityScore, the gain is marginal and comes with a larger decrease in $\mathrm{Recall}_2$. This behavior is consistent with a backbone that benefits from the post-check in some cases but also becomes more exposed to false downgrades when the generated rationale is more variable. The higher decoding temperature used for \texttt{gpt-5-mini} is also consistent with this more mixed downgrade behavior.

For the locally deployed or lower-alignment backbones, the downgrade step becomes less reliable. \texttt{Llama-3.1-70B-Instruct} shows a higher trigger rate and only negligible net benefit, while \texttt{qwen3-30B-A3B-Instruct-2507} and \texttt{deepseek-r1} show small declines in both Macro-F1 and SeverityScore together with larger decreases in $\mathrm{Recall}_2$. These cases indicate that false downgrades can outweigh successful corrections when the generated rationale is less stable or less tightly aligned with the final label.

Appendix Tables~\ref{tab:cm_gpt41} and \ref{tab:cm_deepseek} further illustrate this difference. For \texttt{gpt-4.1}, the confusion matrices show a modest redistribution from class~2 to class~1 after downgrade. Most of this shift corrects previously over-severe predictions on true class~0 and class~1 samples, while only a small number of true class~2 samples are moved downward. This pattern is consistent with the small net gains in Macro-F1 and SeverityScore together with the limited drop in $\mathrm{Recall}_2$. For \texttt{deepseek-r1}, the confusion matrices show a broader reduction in class~2 predictions. Although some of these changes improve classification for true class~0 samples, they also move more true class~1 and class~2 samples downward, which is consistent with the observed drop in $\mathrm{Recall}_2$ and the slight net degradation in Macro-F1 and SeverityScore. The downgrade mechanism should therefore be understood as a lightweight conditional post-check whose benefit depends on rationale quality, rather than as a uniformly beneficial calibration rule.

\subsection{Ablation study}

\begin{figure}[ht]
    \centering
    \includegraphics[width=0.48\textwidth]{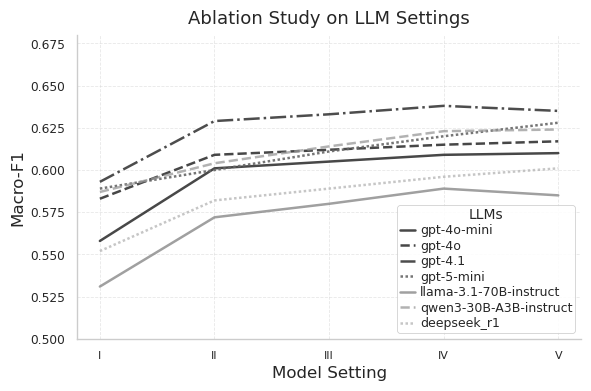}
    \caption{Ablation study of macro-F1 for PDE category prediction across seven LLM backbones. \textbf{I}: target text mode only. \textbf{II}: \textbf{I} + retrieved neighbor reasoning. \textbf{III}: \textbf{II} + conditional free-shots with random-ranking control. \textbf{IV}: \textbf{II} + conditional free-shots with divergence-guided ranking. \textbf{V}: \textbf{IV} + downgrade mechanism.}
    \label{fig:ablation}
\end{figure}

Macro-F1 generally increases as additional context is introduced, and this pattern holds across the evaluated backbones (Figure~\ref{fig:ablation}). Using only the target text mode yields the weakest performance. Adding labeled neighbor reasoning within 1~km improves macro-F1 for every backbone, indicating that geographically proximate analogs and their rationales provide useful local evidence beyond the target description alone. This supports the design role of neighbor retrieval as the primary source of query-time performance gain.

Injecting free-shots under a random-ranking control produces small but generally positive gains over the neighbor-only setting. Replacing this random control with the divergence-guided feature profile yields a further improvement across all evaluated backbones. This pattern provides a compact direct check on the practical contribution of the divergence-informed feature profile. The gain from free-shot injection is not only due to adding more examples, but also depends on how the retained feature profile organizes prototype and hard-boundary selection. The stronger stage-IV results are also consistent with the rationale-grounding role of the divergence-guided design discussed in Sections~\ref{sec:distri} and~\ref{sec:freeshot}.

Enabling the downgrade step has a smaller and more model-dependent effect than neighbor retrieval and divergence-guided free-shot selection. This behavior is also reflected in the downgrade statistics reported in Table~\ref{tab:downgrade_effect}, where the overall effect of the post-check remains limited and model-dependent.

Taken together, the ablation supports the full \textit{R2RAG-Flood} design. Neighbor retrieval provides the main accuracy lift, the random-control comparison shows that divergence-guided feature ranking contributes beyond an arbitrary ordering, and the downgrade step offers a limited model-dependent correction for selected weak-evidence outputs.
\section{Conclusions}

\textit{R2RAG-Flood} is a training-free retrieval-augmented generation framework for post-storm Property Damage Extent (PDE) nowcasting from tabular geospatial predictors. In this framework, labeled flood records are organized into a reasoning-centric knowledge base that stores structured predictors, compact text-mode summaries, and reasoning trajectories. The pipeline combines divergence-informed feature ordering, HUC12-aware retrieval, selected prototypes and hard-boundary cases, and a conservative downgrade post-check to support ordinal PDE prediction without task-specific fine-tuning of the LLM backbone.

In the Harris County Hurricane Harvey case study, \textit{FloodDamageCast*} remains the strongest supervised reference in overall predictive performance. Within the \textit{R2RAG-Flood} family, adding geographically local neighbor reasoning and conditional free-shots improves over a text-mode-only prompt, and the compact ablation comparison further shows that divergence-guided free-shot construction is more effective than a random-ranking control. The downgrade step provides a limited, model-dependent correction for a subset of over-severe predictions. Across the evaluated backbones, the results show that a training-free retrieval pipeline can generate useful PDE predictions together with structured rationales in a realistic flood-damage setting. Under the case-study cost accounting used here, lighter backbones provide stronger severity-per-cost efficiency, whereas stronger backbones offer better overall predictive quality or higher severe-damage recall.

The present study has several limitations. First, the evaluation is restricted to a single Hurricane Harvey case study in Harris County, so the reported results should not be interpreted as evidence of broad geographic generalization. Second, we evaluate a fixed set of LLM backbones under fixed decoding settings, while model behavior may vary under different sampling controls. Third, reasoning quality is assessed using automated metrics under a uniform protocol, which cannot replace domain-expert review. Fourth, the current neighbor ordering is designed to preserve local hydrologic and geographic locality in the present framework; alternative retrieval orders, including feature-similarity-first and hybrid strategies, remain open extensions. Finally, the reported cost comparison is a relative case-study accounting rather than a full lifecycle operational analysis.

Future work will extend the evaluation to additional flood events and regions, test robustness under multiple decoding settings, and incorporate complementary evidence sources such as remote sensing imagery and text damage reports. It will also be valuable to compare the proposed framework with newer supervised classification approaches, including neural dynamic classification, dynamic ensemble learning methods, and fast-learning models such as FEMa, to better understand when retrieval-based reasoning is preferable to purely supervised alternatives. Another practical next step is to introduce domain-expert guidance into text-mode summarization, reasoning templates, exemplar selection, and retrieval design so that prompt context and post-check rules better reflect regional flood characteristics and evolving hazard conditions.


\section*{Code Availability}
The code that supports the findings of this study is available from the corresponding author upon request.

\section*{Data Availability}
The data that support the findings of this study is available from the corresponding author upon request.
\balance
\bibliography{wileyNJD-APA}
\newpage
\appendix
\section{Downgrade Cues}
\label{app:downgrade}
\begin{lstlisting}[language=python,
                    showstringspaces=false,
                    basicstyle=\ttfamily\fontsize{7pt}{8pt}\selectfont,
                    breaklines=true,
                    xleftmargin=0em,
                    xrightmargin=0em,
                    breakindent=0pt,
                    columns=fullflexible,
                    framerule=1pt,
                    resetmargins=true,
                    breakatwhitespace=true
                    ]
SEVERITY_CUES = [
    "deep inundation", "indoor damage", "water entered building",
    "long-lasting flooding", "prolonged", "severe structural",
    "impassable", "major damage", "over-topping", "overtopping",
    "high water depth", "significant damage"
]
LIGHT_CUES = [
    "minor", "shallow", "surface-level", "brief", "quickly receded",
    "passable", "no indoor", "limited", "localized", "light impact",
    "nuisance flooding"
]
UNCERTAIN_CUES = ["uncertain", "insufficient", "not enough evidence", "ambiguous", "unsure"]
\end{lstlisting}
\section{Text Mode Sample}
\label{app:textmode}
Here is the basic tabular information for a training record from Mallard Lake–Cypress Creek.
\begin{lstlisting}[language=python,
                    showstringspaces=false,
                    basicstyle=\ttfamily\fontsize{7pt}{8pt}\selectfont,
                    breaklines=true,
                    xleftmargin=0em,
                    xrightmargin=0em,
                    breakindent=0pt,
                    columns=fullflexible,
                    framerule=1pt,
                    resetmargins=true,
                    breakatwhitespace=true
                    ]
{"index": 429, "zip": "77447", "x": -95.8234621479532, "y": 30.0165830483419, "age": 27.0, "FAR": 0.020766014868859997, "poi_num": 0.0, "Popu_num": 21.871354761343653, "claims_past_50yr": 18, "dis_stream": 24.14419964597232, "elevation": 67.1029411764706, "dis_coa": 73.44553213879476, "hand": 19.75105811403509, "impervious": 10.078431, "roughness": 0.186187946428571, "Poly_num": 39.0, "Rain_max": 13.24, "fndn": 1.661290322580647, "Cluster": "0", "Sum_PDE": 0.045665492259151, "huc12": "120401020103", "PDE_category": 1, "imp_bin": "all", "strata_key": "PDE_category"}
\end{lstlisting}
Each LLM generates its own text mode (4 totally in this study) from the tabular inputs, and due to page limits we only report the text mode produced by \texttt{gpt-4.1}.
\begin{lstlisting}[language=python,
                    showstringspaces=false,
                    basicstyle=\ttfamily\fontsize{7pt}{8pt}\selectfont,
                    breaklines=true,
                    xleftmargin=0em,
                    xrightmargin=0em,
                    breakindent=0pt,
                    columns=fullflexible,
                    framerule=1pt,
                    resetmargins=true,
                    breakatwhitespace=true
                    ]
{'text_mode_gpt-4.1': 'Poly_num is 39, Popu_num is about 21.87. The fndn is 1.66 ft, hand measures 19.75 m, and Rain_max is 13.24 inches. Elevation averages 67.1 ft, dis_coa is 73.45 km. There have been 18 claims_past_50yr. Roughness is 0.19, FAR is 0.021, and impervious surface is 10.08%. The average age is 27 years, and dis_stream is 24.14 km.'}
\end{lstlisting}

\section{Reasoning Trajectory Sample}
The corresponding reasoning trajectory for the record shown in Appendix~\ref{app:textmode} is as follows:
\label{app:train}
\begin{lstlisting}[language=python,
                    showstringspaces=false,
                    basicstyle=\ttfamily\fontsize{7pt}{8pt}\selectfont,
                    breaklines=true,
                    xleftmargin=0em,
                    xrightmargin=0em,
                    breakindent=0pt,
                    columns=fullflexible,
                    framerule=1pt,
                    resetmargins=true,
                    breakatwhitespace=true
                    ]

{'reasoning_gpt-4.1': '<think>For occurrence, the foundation is moderate (1.66 ft), and HAND is very high (19.75 m), which is highly protective. Elevation is also high (67.1 ft). However, there are 18 claims in the past 50 years, indicating repeated flood impacts. Rainfall is significant (13.24 inches), and impervious surface is moderate (10.08%), with a low FAR (0.021). The distance to stream (24.14 km) reduces direct fluvial risk. Occurrence resolves to 1, as the strong history of claims outweighs the robust elevation and HAND protections. For severity, while rainfall is notable, the high HAND and elevation, moderate imperviousness, and low FAR limit the likelihood of extreme losses. Severity resolves to 1. Despite the strong protective elevation and HAND, the repeated claims history justifies a non-zero occurrence but not escalation to high severity. Based on these factors, it is reasonable to claim PDE_category is 1.</think><answer>1</answer>'}
\end{lstlisting}
\section{Free-shots Sample}
\label{app:freeshot}
Because Mallard Lake–Cypress Creek (sample used in Appendix~\ref{app:textmode}) contains fewer than 100 records, we instead display the global free-shots that provide class prototypes and hard examples.
\begin{lstlisting}[language=python,
                    showstringspaces=false,
                    basicstyle=\ttfamily\fontsize{7pt}{8pt}\selectfont,
                    breaklines=true,
                    xleftmargin=0em,
                    xrightmargin=0em,
                    breakindent=0pt,
                    columns=fullflexible,
                    framerule=1pt,
                    resetmargins=true,
                    breakatwhitespace=true
                    ]

"prototypes": {
    "0": [
      {
        "row_pos": 263,
        "PDE_category": 0,
        "features": {
          "feature": value,
        },
        "weighted_zdist": 0.18196343236392964,
        "per_feature": {
          "feature": zdist
        },
        "why_selected": "Selected as class 0 prototype by minimal weighted z-distance; d=0.1820. Top contributors: FAR: 0.429, impervious: 0.386, elevation: 0.331.",
        "text_mode": text_mode,
        "reasoning": training_reasoning,
      },
      {...}
    ],
    "1": [...],
    "2": [...]
},
"hard_examples": {
    "occurrence_boundary": {
      "for_0": {
        "row_pos": 319,
        "PDE_category": 0,
        "features": {
          "feature": value,
        },
        "why_selected": "Closest to 0/1 occurrence boundary (margin=0.0006; d0=0.4455, d12=0.4461).",
        "text_mode": text_mode,
        "reasoning": training_reasoning,
      },
      "for_1": {...}
    },
    "severity_boundary": {
      "for_2": {
        "row_pos": 665,
        "PDE_category": 2,
        "features": {
          "feature": value,
        },
        "why_selected": "Closest to 1/2 severity boundary (margin=0.0013; d1=0.8826, d2=0.8812).",
        "text_mode": text_mode,
        "reasoning": training_reasoning,
      }
    }
}
    
\end{lstlisting}

\section{Prediction Reasoning Sample}
We present a correctly predicted test record from Willow Creek. Its tabular attributes, converted text mode, predicted label, and associated \texttt{gpt-4.1} predicted reasoning are shown as follows.
\label{app:pred}
\begin{lstlisting}[language=python,
                    showstringspaces=false,
                    basicstyle=\ttfamily\fontsize{7pt}{8pt}\selectfont,
                    breaklines=true,
                    xleftmargin=0em,
                    xrightmargin=0em,
                    breakindent=0pt,
                    columns=fullflexible,
                    framerule=1pt,
                    resetmargins=true,
                    breakatwhitespace=true
                    ]
{"index": 5448, "zip": "77389", "x": -95.50207982520224, "y": 30.106515152470354, "age": 30.0, "FAR": 0.12494365122514271, "poi_num": 1.0, "Popu_num": 593.229423749367, "claims_past_50yr": 26, "dis_stream": 4.273092975385168, "elevation": 47.25, "dis_coa": 50.1185255739239, "hand": 14.415917024704619, "impervious": 30.820312, "roughness": 0.103759821428571, "Poly_num": 146.0, "Rain_max": 15.04, "fndn": 1.566091954022989, "Cluster": "0", "Sum_PDE": 0.453263300885126, "huc12": "120401020210", "PDE_category": 1, "imp_bin": "all", "strata_key": "PDE_category", "pred_PDE_category_gpt-4.1": 1}

\end{lstlisting}
\begin{lstlisting}[language=python,
                    showstringspaces=false,
                    basicstyle=\ttfamily\fontsize{7pt}{8pt}\selectfont,
                    breaklines=true,
                    xleftmargin=0em,
                    xrightmargin=0em,
                    breakindent=0pt,
                    columns=fullflexible,
                    framerule=1pt,
                    resetmargins=true,
                    breakatwhitespace=true
                    ]

{'text_mode_gpt-4.1': 'Poly_num 146, Popu_num about 593 people, foundation height 1.57 ft, HAND 14.4 m, Rain_max 15.04 inches, elevation 47.3 ft, 50.12 km from coast, 26 insurance claims in 50 years, roughness 0.10, FAR 0.125, 1 point of interest, impervious surface 30.82%, average building age 30 years, 4.27 km to nearest stream.',}
\end{lstlisting}
\begin{lstlisting}[language=python,
                    showstringspaces=false,
                    basicstyle=\ttfamily\fontsize{7pt}{8pt}\selectfont,
                    breaklines=true,
                    xleftmargin=0em,
                    xrightmargin=0em,
                    breakindent=0pt,
                    columns=fullflexible,
                    framerule=1pt,
                    resetmargins=true,
                    breakatwhitespace=true
                    ]
{"pred_PDE_category_gpt-4.1": 1, 
 "pred_reasoning_gpt-4.1": "<think>Target shows several opposing signals: relatively large population (~593), nontrivial FAR (~0.125) and impervious (~30.8%) increase exposure and runoff potential; foundation elevation (~1.57 ft) offers modest selfprotection but not deep clearance. HAND is high (~14.4 m) which is protective, but the distance to nearest stream is short (~4.27 km), increasing possible fluvial or routed overland influence; max daily rain is extreme (~15.0 in) and there are 26 historical claims. Neighbors are split: neighbor#1 (0.499 km) labels 1 emphasizing high HAND but substantial claims/population and closeish stream merit nonzero occurrence; neighbor#2 (0.499 km) labels 2 because of denser built environment and closer stream in that exemplar; neighbor#3 (0.5 km) labels 0 citing very high HAND and buffering. I weigh the two closest neighbors first and consider target siting: substantial asset base and moderate FAR plus short stream distance but also strong HAND. That mixture aligns with a medium (1) PDE assignment in the prototypes: nonzero occurrence possible but not necessarily catastrophic across the cell. Therefore predict medium (1).</think><answer>1</answer>"}
\end{lstlisting}

\section{Illustrative Confusion Matrices With and Without Downgrade}
To provide a compact view of the class-level effect of the downgrade step, we report illustrative confusion matrices for one strong overall backbone (\texttt{gpt-4.1}) and one lower-alignment backbone (\texttt{deepseek-r1}). Rows denote true labels and columns denote predicted labels.

\begin{table}[h]
\centering
\caption{Confusion matrices for \texttt{gpt-4.1} without and with downgrade.}
\label{tab:cm_gpt41}
\setlength{\tabcolsep}{4pt}
\renewcommand{\arraystretch}{1.0}
\scriptsize
\begin{tabular}{c|ccc|ccc}
\toprule
& \multicolumn{3}{c|}{Without downgrade} & \multicolumn{3}{c}{With downgrade} \\
True $\backslash$ Pred & 0 & 1 & 2 & 0 & 1 & 2 \\
\midrule
0 & 846 & 830 & 201 & 846 & 900 & 131 \\
1 & 50  & 1923 & 90  & 50  & 1939 & 74  \\
2 & 80  & 292  & 302 & 80  & 298  & 296 \\
\bottomrule
\end{tabular}
\end{table}

\begin{table}[hb]
\centering
\caption{Confusion matrices for \texttt{deepseek-r1} without and with downgrade.}
\label{tab:cm_deepseek}
\setlength{\tabcolsep}{4pt}
\renewcommand{\arraystretch}{1.0}
\scriptsize
\begin{tabular}{c|ccc|ccc}
\toprule
& \multicolumn{3}{c|}{Without downgrade} & \multicolumn{3}{c}{With downgrade} \\
True $\backslash$ Pred & 0 & 1 & 2 & 0 & 1 & 2 \\
\midrule
0 & 865 & 765 & 247 & 900 & 820 & 157 \\
1 & 80  & 1789 & 194 & 130 & 1764 & 169 \\
2 & 55  & 290  & 329 & 70  & 295  & 309 \\
\bottomrule
\end{tabular}
\end{table}
\end{document}